\documentclass[10pt,twocolumn]{article}
\usepackage[letterpaper,top=.76in,bottom=.78in,left=.73in,right=.73in]{geometry}
\usepackage[T1]{fontenc}
\usepackage{times,microtype}
\usepackage{amsmath,amssymb,amsfonts,amsthm}
\usepackage{graphicx,booktabs,tabularx,multirow,makecell,xcolor}
\usepackage{algorithm,algorithmic}
\usepackage[font=small,labelfont=bf]{caption}
\usepackage[breaklinks=true,hidelinks]{hyperref}
\newcommand{\method}{\textsc{ReliFlow}}
\newcommand{\addraw}{A_{\mathrm{raw}}}
\newcommand{\addcal}{A_{\mathrm{cal}}}
\newcommand{\expect}{\mathbb E}
\newcommand{\vct}[1]{\boldsymbol{#1}}
\title{\vspace{-1em}\bfseries ReliFlow: Reliability-Calibrated Residual Transport for Combinatorial Single-Cell Perturbation Prediction}
\author{Jinrui David$^{1}$ \qquad Yun Peng$^{1}$\\[.2em]
\normalsize $^{1}$Fudan University, Shanghai, China}
\date{}
\begin{document}
\twocolumn[\maketitle\vspace{-1.6em}
\begin{abstract}
Predicting the transcriptional response to combined perturbations requires distinguishing interactions between interventions from errors in their individual effect estimates. The latter become more pronounced when single-perturbation evidence is sparse or withheld. We introduce \method{}, a conditional flow-matching model that estimates the reliability of single-gene effects from training-only cross-fitting and uses those estimates to calibrate the additive source distribution. The velocity is decomposed into a known correction of the additive reference and a learned residual transport field. On Norman genetic perturbations, \method{} reduces mean absolute error (MAE) from 0.142 to 0.131 for unseen combinations of observed singles and from 0.198 to 0.178 when constituent singles are withheld. On held-out ComboSciPlex drug combinations, MAE decreases from 0.191 to 0.176. Non-additive residual correlation improves from 0.574 to 0.638, and component ablations identify the contribution of reliability weighting. The results indicate that accounting for uncertain constituent effects improves generalization to unseen genetic and chemical combinations, with a moderate increase in training cost.
\end{abstract}\vspace{.35em}]

\section{Introduction}
A transcriptional response to a genetic or chemical perturbation is a distribution over heterogeneous cellular states rather than a single deterministic vector. Predicting that distribution for interventions not yet measured could help prioritize experiments and explore regulatory interactions, but the space of potential combinations quickly exceeds the available observations~\cite{norman2019exploring,roohani2024predicting,lotfollahi2023predicting}. A typical prediction system can learn common responses from training perturbations while missing rare response states or interaction-specific changes. The central difficulty is therefore not only to reconstruct the dominant expression pattern, but also to determine which evidence supports the predicted deviation from a reference state.

A natural reference is additive composition. When the effects of interventions $a$ and $b$ are measured separately, their summed expression shifts offer a useful prediction of the combined treatment. Indeed, simple linear strategies remain competitive against considerably larger perturbation models on some benchmarks~\cite{ahlmann2025deep,csendes2025benchmarking}. The value of additive composition, however, depends on how accurately its constituent effects were estimated. This is particularly consequential when evaluating generalization beyond familiar intervention combinations. For a pair of well-characterized perturbations, the additive component may be stable and the residual largely reflects interactions. For a pair involving an unseen perturbation, the residual also contains error inherited from extrapolating a single-effect predictor. Treating both cases as equivalent may obscure what the model is learning.

Conditional flow matching provides a flexible way to learn distributions without assuming a pairing between individual control and treated cells~\cite{lipman2022flow,bunne2023learning}. An additive estimate can instead be used as the flow source, leaving a conditional velocity field to learn the residual transport. This is an informative starting point: it reduces the burden of generating a complete expression vector from an unrelated noise distribution. Yet an additive anchor estimated from sparse evidence is not equally trustworthy across genes and perturbations. If its errors are ignored, a velocity model may use its capacity to repair the anchor rather than model higher-order biological effects. The issue is most acute precisely where predictive models are expected to extrapolate.

\textbf{We ask whether the reliability of observed single-perturbation evidence should determine where residual transport begins and which part of the required correction is learned.} We develop \method{} to make this choice explicit. It estimates a gene-wise support scale using cross-fitted training perturbations, shrinks uncertain additive shifts toward a learned prior, and adjusts source variability accordingly. We then decompose the conditional velocity into a known correction that restores the raw additive reference and a learned field that targets the remaining interaction-dependent transport. In this way, anchor uncertainty is represented rather than silently mixed into the nonlinear residual.

The model makes three distinctions operational. First, reliability is estimated from training data and transferred to excluded interventions through their representations; it cannot use the held-out outcomes that it seeks to predict. Second, full-response reconstruction and recovery of non-additive effects receive separate diagnostic analyses. Third, improvements must be assessed on matched partitions and compute budgets rather than inferred from attractive samples or one aggregate score. These choices reflect a broader principle in predictive modeling: model confidence and downstream utility must be tied to the evidence actually available, not simply to the plausibility of generated outputs~\cite{xi2025uncertainty}.

We evaluate the resulting model using condition-disjoint genetic and chemical perturbation splits, interaction-sensitive metrics, and component ablations.

\section{Related work}
\subsection{Learning perturbation response distributions}
Single-cell response prediction has progressed from latent perturbation arithmetic in scGen~\cite{lotfollahi2019scgen} to neural optimal transport~\cite{bunne2023learning}, compositional perturbation autoencoders~\cite{lotfollahi2023predicting}, graph-based gene perturbation models such as GEARS~\cite{roohani2024predicting}, and predictive systems for both chemical and genetic interventions~\cite{qi2024predicting,yu2025perturbnet}. Drug combinations can be assayed through combinatorial transcriptomic screens~\cite{mathur2022combi}, while statistical models such as GPerturb offer an alternative account of gene-level variation~\cite{xing2025gperturb}. Recent models target the response distribution itself, including direct flow models and conditional distribution shift prediction~\cite{yu2026scdfm,klein2025cellflow,ramakrishnan2025modeling}. Heterogeneous outcomes and multi-response populations make it important to evaluate more than mean error~\cite{song2025decoding,mejia2025diversity}.

Large-scale pretrained representations offer reusable gene features. These include scBERT~\cite{yang2022scbert}, transformer-based cell representations~\cite{khan2023reusability}, scGPT~\cite{cui2024scgpt}, Geneformer~\cite{theodoris2023transfer}, and universal cell embeddings~\cite{rosen2023universal}. The deep generative scVI framework models technical variation and uncertainty in single-cell expression~\cite{lopez2018scvi}; foundation-model approaches and virtual-cell benchmarks extend the scope of perturbation prediction~\cite{adduri2025predicting,roohani2025virtual}. Benchmarking studies report competitive performance from strong linear baselines and sensitivity to evaluation metrics and data composition~\cite{ahlmann2025deep,csendes2025benchmarking,vinas2025systema}. These observations motivate separating an uncertain single-effect estimate from non-additive response generation.

\subsection{Transport, generative dynamics and evidence quality}
Diffusion modeling~\cite{ho2020ddpm}, score-based stochastic dynamics~\cite{song2021sde}, and flow matching~\cite{lipman2022flow} provide related approaches to learning complex distributions. Stochastic interpolants formalize the relationship between diffusion and deterministic transport~\cite{albergo2025interpolants}. Kernel two-sample comparisons furnish distribution-level losses without observed cell-to-cell correspondence~\cite{gretton2006kernel,gretton2012test}. This is important because destructive expression profiling does not provide paired before-after observations for the same cell. Structure priors can be sensitive to dropout and nuisance variation~\cite{dai2024gene,zhou2025quantifying}. Differential attention and other selective representation mechanisms provide architectural options but do not themselves establish calibrated perturbation effects~\cite{ye2025differential}.

Reliability and grounding also arise in other forms of conditional prediction. Structured multimodal diffusion has been studied for multi-agent trajectory forecasting~\cite{yang2026unified}, whereas radiology grounding and radiographic world models connect generated representations to evidence in clinical imaging~\cite{xi2026grounding,xi2026radiographic}. Medical visual reasoning and uncertainty-aware imaging models provide further perspectives on distinguishing plausible outputs from supported predictions~\cite{xi2026medlvr,xi2025uncertainty}. These studies address different tasks rather than direct single-cell baselines. Here we take the narrower position that training-supported single-effect reliability should explicitly modulate the source distribution used for conditional transport.

\section{Problem setting}
Let $c\in\mathbb R^G$ denote expression from a control cell over $G$ genes, and let $p=\{j_1,\ldots,j_m\}$ denote an intervention set. For each condition $p$, we observe independent samples from a treated-cell distribution $P_p$. We also have a control population $P_0$. Because cell measurements are destructive, $c\sim P_0$ and $x\sim P_p$ are \emph{not} matched observations of the same biological cell. We use $(c,p,x)$ only as a conditional transport training tuple, never as an observed counterfactual pair.

Suppose an effect decoder $d_\phi(c,e_j)$ predicts the response shift for single intervention $j$, where $e_j$ is a perturbation representation. The raw additive estimate is
\begin{equation}
\addraw(c,p)=c+\sum_{j\in p}d_\phi(c,e_j).
\label{eq:raw}
\end{equation}
If all component effects were perfectly estimated and acted independently, this reference would be sufficient. For an observed treated profile $x$, define a \emph{reference-relative} residual $s_{\mathrm{raw}}=x-\addraw(c,p)$. This residual is a useful transport target, but because $(c,x)$ are unpaired it should not be interpreted as an identified individual-cell causal interaction. At the population level, departures from additive mean response can be estimated from separate treated and control samples, provided that all component perturbation conditions are observed.

We distinguish two evaluation regimes. \emph{Composition} excludes combinations of training-observed singles; \emph{extrapolation} excludes one or more constituent singles and combinations containing them. The latter requires inference about both the single-effect predictions and their interactions. We keep these regimes separate throughout the analysis.

\section{Method: Reliability-calibrated transport}
\label{sec:method}
\begin{figure*}[t]
\centering\includegraphics[width=.98\textwidth]{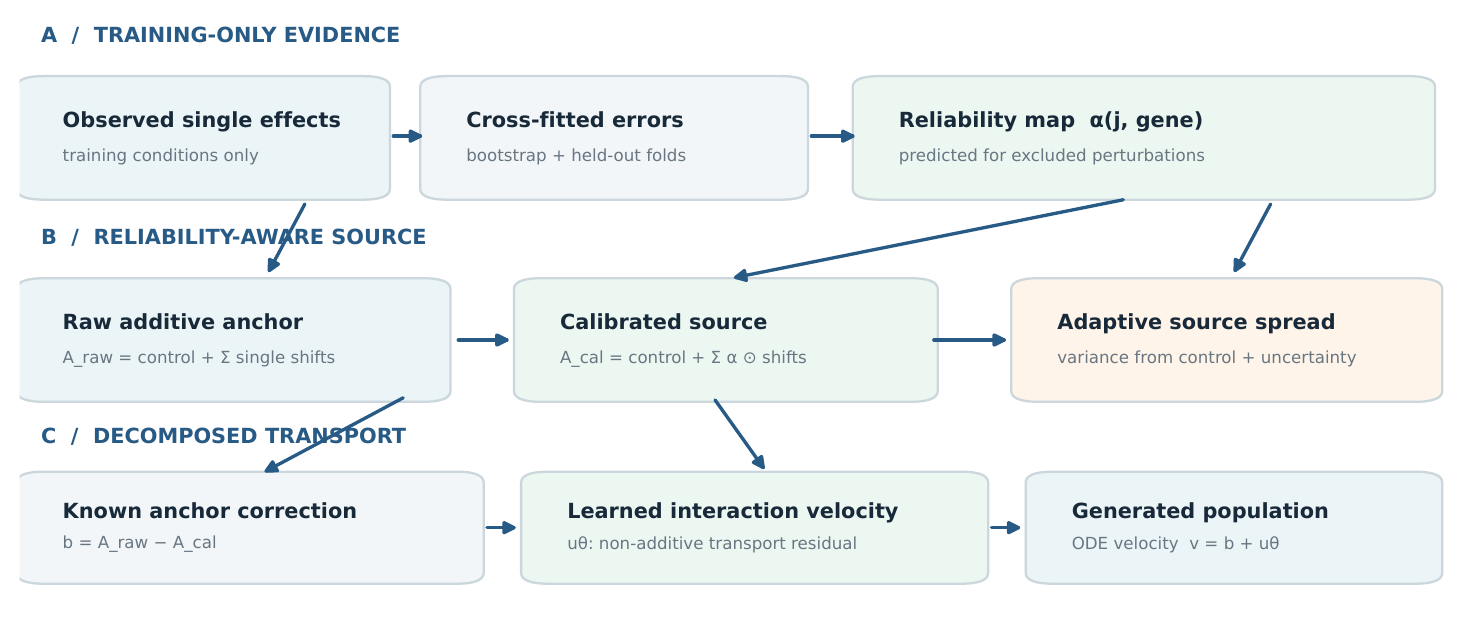}
\caption{\textbf{\method{} architecture.} Training-only single-perturbation evidence is used to estimate gene-wise reliability. The resulting reliability map modifies the additive source and its stochastic spread. The full transport field is decomposed into a deterministic correction from the calibrated to the raw additive anchor and a learned interaction-residual velocity. The architecture separates evidence calibration, anchor construction, and interaction-conditioned velocity learning.}\label{fig:architecture}
\end{figure*}

\subsection{Estimating reliability without test leakage}
The single-effect decoder $d_\phi$ is trained using only the training split. To evaluate the credibility of each estimated gene response, we cross-fit it across training conditions. For an observed training single $j$, the measured population-average effect on gene $g$ is
\begin{equation}
\widehat\Delta_{jg}=\overline x_{j,g}-\overline c_g,
\label{eq:obs_delta}
\end{equation}
with both means computed from the relevant training populations. We divide the training perturbation identities into $K$ folds and let $f(j)$ denote the fold containing $j$. The cross-fitted effect $\widehat\Delta_{jg}^{(-f(j))}$ is predicted by a model trained without response profiles from any perturbation in that fold. This leave-condition-out design requires $e_j$ to be computable for an excluded gene identity from shared gene features, rather than an embedding trained only as a separate lookup parameter. We define an error-scale target
\begin{equation}
\widehat u^2_{jg}=\big(\widehat\Delta^{(-f(j))}_{jg}-\widehat\Delta_{jg}\big)^2
+\widehat{\mathrm{Var}}_{\mathrm{boot}}(\widehat\Delta_{jg}),
\label{eq:scale_target}
\end{equation}
where the bootstrap term captures sampling variation in estimating the treated-control shift. This is a training-data reliability proxy, not a calibrated posterior variance. Cross-fitting prevents the same response observations from simultaneously fitting the effect decoder and assessing its error.

A nonnegative reliability head $u_\psi(e_j,c)_g$ predicts these scales from the perturbation representation and control context. We train it using a robust loss on log scales,
\begin{equation}
\mathcal L_{\mathrm{rel}}=
\expect_{j\in\mathcal P_{\mathrm{single}}^{\mathrm{train}},g}
\rho\!\left(\log(\varepsilon+u_\psi^2)-\log(\varepsilon+\widehat u^2_{jg})\right),
\label{eq:rel_loss}
\end{equation}
where $\rho$ is a Huber penalty. At test time the network produces $u_\psi$ without observing test treatment outcomes. For an excluded perturbation, its scale is therefore an extrapolated quantity; poor scale extrapolation is an explicit failure mode to evaluate.

\subsection{A support-calibrated additive source}
Let $v_g>0$ be the empirical across-perturbation variance of training single-effect means for gene $g$, with a variance floor for stability. We define a reliability weight
\begin{equation}
\alpha_{jg}(c)=\frac{v_g}{v_g+u_\psi^2(e_j,c)_g+\varepsilon},
\qquad 0<\alpha_{jg}\leq1.
\label{eq:weight}
\end{equation}
This is a deterministic shrinkage rule, not a claim of optimal Bayesian estimation. It expresses a testable bias: when the estimated single effect is unreliable relative to the usual variability of training effects, the model should rely less on that predicted shift when choosing a starting point. The calibrated anchor becomes
\begin{equation}
\addcal(c,p)=c+\sum_{j\in p}\vct\alpha_j(c)\odot d_\phi(c,e_j).
\label{eq:cal_anchor}
\end{equation}
We use a gene-wise stochastic source
\begin{equation}
\begin{split}
 z_0 &= \addcal(c,p)+\epsilon,\\
 \epsilon &\sim\mathcal N\!\left(0,\operatorname{diag}(\vct\sigma_0^2(c,p))\right),\\
 \sigma^2_{0,g}&=\tau^2\sigma^2_{\mathrm{ctrl},g}
 +\kappa^2\sum_{j\in p}\alpha_{jg}^2(c)u_{\psi}^2(e_j,c)_g.
\end{split}
\label{eq:source}
\end{equation}
The first term preserves control-population variability; the second allows uncertainty in estimated intervention effects to affect the initial source distribution. Coefficients $\tau$ and $\kappa$ are chosen on validation data only. An unreliable additive prediction is thus neither accepted without qualification nor replaced by a universal noise source.

\subsection{Separating anchor repair from learned interaction}
The crucial structural choice is to keep track of the difference between the calibrated and raw additive anchors:
\begin{equation}
 b(c,p)=\addraw(c,p)-\addcal(c,p).
\label{eq:known_correction}
\end{equation}
This correction is computable at inference time. Given $z_0$ from Eq.~\eqref{eq:source} and an independently sampled treated profile $x$, define the straight conditional path
\begin{equation}
 x_t=(1-t)z_0+tx,\qquad t\sim\mathcal U(0,1).
\label{eq:path}
\end{equation}
Its conditional target velocity has an exact algebraic decomposition:
\begin{equation}
\underbrace{x-z_0}_{v^*}
 =\underbrace{(\addraw-\addcal)}_{b}
 +\underbrace{(x-\addraw-\epsilon)}_{u^*}.
\label{eq:decomposition}
\end{equation}
We parameterize the full velocity as
\begin{equation}
 v_\theta(x_t,t,c,p)=b(c,p)
 +u_\theta(x_t,t,c,p,\vct\alpha_p).
\label{eq:velocity}
\end{equation}
The learned field $u_\theta$ therefore receives the reference-relative target $u^*=x-\addraw-\epsilon$, while the anchor-repair term $b$ is represented analytically. This prevents a learned network from having to rediscover a correction already known from the two anchors. It does not, however, make the learned residual an identified biological mechanism: nuisance variation and unpaired sampling remain part of the distributional problem.

\begin{figure}[t]
\centering\includegraphics[width=\columnwidth]{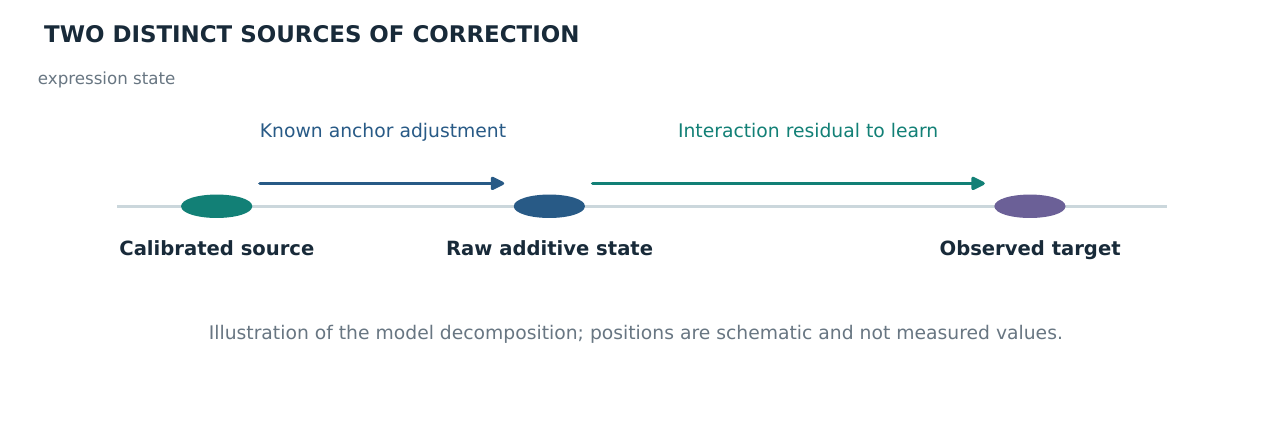}
\caption{\textbf{Transport decomposition.} The known vector $b$ bridges the reliability-calibrated source and raw additive estimate, while $u_\theta$ learns the remaining transport. Diagrammatic distances are not empirical measurements.}\label{fig:decomp}
\end{figure}

\subsection{Interaction-conditioned velocity network}
To represent conditional effects, a gene-token encoder receives control expression, gene identity, and a perturbation embedding. The encoder uses sparse gene-interaction attention; reliability weights are supplied as conditioning signals rather than conflated with edge attention. The input token for gene $g$ is
\begin{equation}
 \begin{aligned}
h_g^0={}&E_g(g)+E_x(x_{t,g})+E_c(c_g)\\
&+E_r(\alpha_{p,g},\sigma_{0,g})+E_p(p)+E_t(t).
\end{aligned}
\label{eq:tokens}
\end{equation}
A transformer or sparse graph-attention stack maps $(h_g^0)_g$ to $u_\theta$. Here $\alpha_{p,g}=|p|^{-1}\sum_{j\in p}\alpha_{jg}$ summarizes reliability for a multi-perturbation condition. The term $E_r$ is important: two examples with identical nominal interventions may warrant different reliance on their additive anchors when training coverage or predicted error scales differ. We avoid imposing an attention graph estimated using test conditions; any graph prior must be learned from training controls or training perturbations only.

\subsection{Distributional and interaction constraints}
The fundamental objective is the conditional flow-matching loss
\begin{equation}
\begin{aligned}
\mathcal L_{\mathrm{FM}}=\expect\,\frac1G\big\|&u_\theta(x_t,t,c,p,\vct\alpha_p)\\
&-(x-\addraw(c,p)-\epsilon)\big\|_2^2.
\end{aligned}
\label{eq:fm}
\end{equation}
For practical population supervision we use the intermediate-time endpoint estimator
\begin{equation}
 \widehat x_1=x_t+(1-t)\big[b(c,p)+u_\theta(x_t,t,c,p,\vct\alpha_p)\big].
\label{eq:endpoint}
\end{equation}
Samples are grouped by perturbation condition before estimating population discrepancies. We use a Gaussian-mixture kernel to match the generated and observed condition-specific endpoint populations,
\begin{equation}
\mathcal L_{\mathrm{pop}}=\operatorname{MMD}^2(\widehat X_{1,p},X_{1,p}).
\label{eq:mmd}
\end{equation}
Because the kernel is translation-invariant, subtracting the same fixed additive reference from both populations leaves this MMD unchanged. We therefore use one population discrepancy rather than presenting two equivalent shifted losses as independent constraints~\cite{gretton2006kernel,gretton2012test}.

Where training contains the single effects of $a$ and $b$ and the observed double $(a,b)$, their mean non-additive response can also be estimated directly:
\begin{equation}
 m_{ab}^{\mathrm{obs}}=\overline x_{ab}-\overline x_a-
 \overline x_b+\overline c.
\label{eq:synergy_obs}
\end{equation}
Let $\widehat m_{ab}=\expect_c[\widehat x_{ab}-\addraw(c,\{a,b\})]$ be the generated reference-relative mean residual. On \emph{training-observed} doubles only, the auxiliary moment constraint is
\begin{equation}
\mathcal L_{\mathrm{pair}}=
\expect_{(a,b)\in\mathcal P_{\mathrm{double}}^{\mathrm{train}}}
\frac1G\|\widehat m_{ab}-m_{ab}^{\mathrm{obs}}\|_2^2.
\label{eq:pair_loss}
\end{equation}
We explicitly exclude held-out combinations from this term. Otherwise an apparently strong test performance could result from leaking the mean response of the target condition into training.

The complete training objective is
\begin{equation}
\begin{split}
\mathcal L={}&\mathcal L_{\mathrm{FM}}+
\lambda_{\mathrm{pop}}\mathcal L_{\mathrm{pop}}+
\lambda_{\mathrm{pair}}\mathcal L_{\mathrm{pair}}\\
&+\lambda_{\mathrm{rel}}\mathcal L_{\mathrm{rel}}+
\lambda_{\mathrm{single}}\mathcal L_{\mathrm{single}},
\end{split}
\label{eq:total}
\end{equation}
where $\mathcal L_{\mathrm{single}}$ is the training-only single-effect decoder loss. Its population-based implementation is given in Appendix~\ref{app:single}. All coefficients must be selected without consulting test scores. The loss terms jointly constrain reliability estimation, condition-level response distributions, and the residual transport.

\subsection{Training and generation}
Reliability-scale fitting is separated from full-model training. Cross-fitted errors are computed only on training perturbations. The support network $u_\psi$ is then fitted on these error targets; finally, the calibrated source and interaction field are optimized using the training conditions. At inference, no observed treated population is required. For a new $p$, the model predicts $d_\phi$, $u_\psi$, $\vct\alpha_p$, and $z_0$, then numerically integrates
\begin{equation}
 \frac{dX_t}{dt}=b(c,p)+u_\theta(X_t,t,c,p,\vct\alpha_p),
 \quad X_0=z_0.
\label{eq:ode}
\end{equation}
The generated condition-specific population comprises independent trajectories from a batch of control cells and source noise samples. The source spread is not automatically a confidence interval for the final output; empirical calibration must be tested separately.

\section{Experiments}
\label{sec:exp}

\subsection{Data, preprocessing, and held-out partitions}
\begin{figure*}[t]
\centering\includegraphics[width=.82\textwidth]{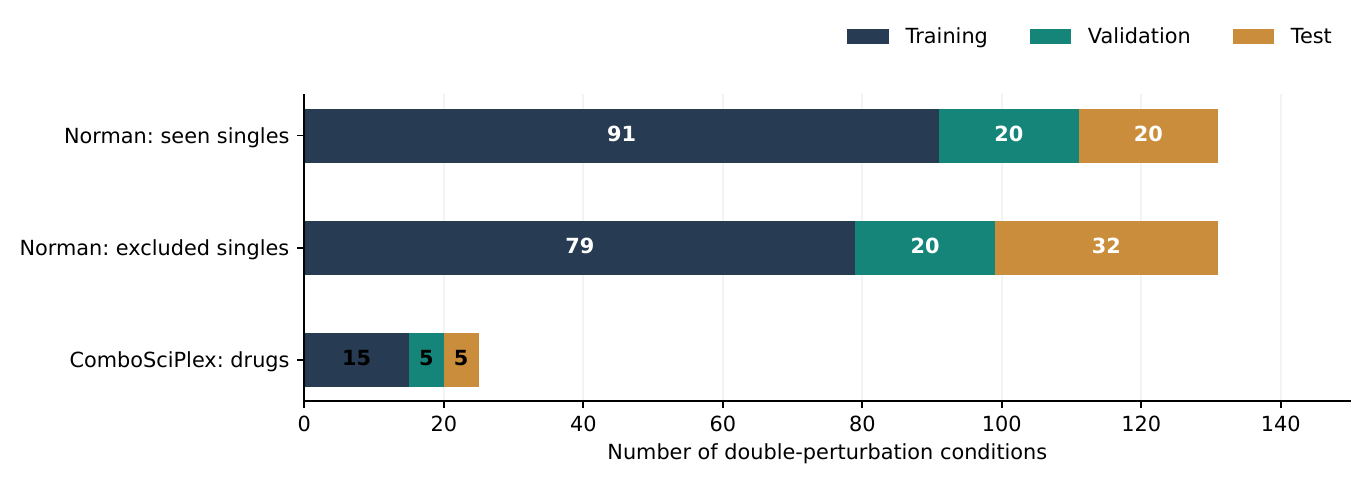}
\caption{\textbf{Double-perturbation splits.} Counts are numbers of distinct combination conditions, not cells. Single-perturbation counts and exact cell counts are given in Tables~\ref{tab:norman_split} and~\ref{tab:drug_split}. In the excluded-single setting, twelve single-gene perturbations and combinations containing the excluded genes are reserved for testing.}
\label{fig:splits}
\end{figure*}

We evaluated genetic perturbation prediction on the Norman dataset~\cite{norman2019exploring} and chemical combination prediction on ComboSciPlex~\cite{mathur2022combi,srivatsan2020massively}. We split perturbation conditions, rather than individual cells, before fitting predictors or selecting hyperparameters. Expression was normalized by counts per 10,000 followed by $\log(1+x)$. The 5,000 highly variable genes (HVGs) were selected using training cells only. We evaluated three random seeds in each dataset. Figure~\ref{fig:splits} shows the allocation of observed double-perturbation conditions.

\begin{table*}[t]
\centering\small
\caption{\textbf{Norman genetic perturbation evaluation.} Exact counts for the two fixed partition schemes. The seen-single test comprises unseen combinations of interventions whose single effects are available during training. The excluded-single test additionally withholds twelve constituent perturbations and their related doubles.}
\label{tab:norman_split}
\begin{tabular}{lcc@{\hspace{12pt}}lcc}
\toprule
& Seen singles & Excluded singles & & Seen singles & Excluded singles \\
\midrule
Training single conditions & 105 & 93 & Training double conditions & 91 & 79 \\
Validation single conditions & 0 & 0 & Validation double conditions & 20 & 20 \\
Test single conditions & 0 & 12 & Test double conditions & 20 & 32 \\
Training cells & 68,000 & 62,000 & Validation cells & 10,000 & 9,000 \\
Test cells & 13,000 & 20,000 & Selected genes & 5,000 HVGs & 5,000 HVGs \\
Split unit & Condition & Perturbation identity & Random seeds & 3 & 3 \\
\bottomrule
\end{tabular}
\vspace{2pt}
\end{table*}

The Norman seen-single partition contains 105 training singles and 91/20/20 training/validation/test double conditions, with 68,000/10,000/13,000 cells, respectively. In the excluded-single partition, training uses 93 singles and 79 double conditions, validation uses 20 doubles, and testing uses 12 withheld singles and 32 doubles; the corresponding cell counts are 62,000/9,000/20,000. There are no validation singles in either partition. The excluded genes occur in no training singles, training doubles, or validation conditions. All these counts are exact.

\begin{table}[t]
\centering\small
\caption{\textbf{ComboSciPlex evaluation.} Each held-out pair is disjoint from training and from the other evaluation split.}
\label{tab:drug_split}
\begin{tabular}{lr}
\toprule
Setting & Value \\
\midrule
Cell line & A549 \\
Measured single-drug conditions & 7 \\
Measured double-drug conditions & 25 \\
Training single-drug conditions & 7 \\
Training double-drug conditions & 15 \\
Validation double-drug conditions & 5 \\
Test double-drug conditions & 5 \\
Training / validation cells & 43,000 / 9,000 \\
Test cells & 11,378 \\
Genes & 5,000 \\
Random seeds & 3 \\
\bottomrule
\end{tabular}
\end{table}

ComboSciPlex comprises A549 responses to seven single-drug conditions and 25 drug pairs. All seven measured singles and 15 double-drug conditions are in training; five double conditions are used for validation and five for testing (Table~\ref{tab:drug_split}). The split contains 43,000 training cells, 9,000 validation cells, and 11,378 test cells. The complete identities of the five validation and five test pairs appear in Appendix~\ref{app:drug_conditions}. Both evaluation splits are condition-disjoint and use the same 5,000-gene normalization procedure as Norman.

\subsection{Baselines, metrics, and experimental controls}
We compared \method{} with the additive predictor, CPA~\cite{lotfollahi2023predicting}, GEARS~\cite{roohani2024predicting}, scDFM~\cite{yu2026scdfm}, and CellFlow~\cite{klein2025cellflow} where applicable. Each tabulated comparison uses the corresponding partition defined above. Mean absolute error (MAE) and mean squared error (MSE) measure expression reconstruction; differential-expression Spearman correlation (DE-Spearman or DE-$\rho$) measures gene-rank agreement; effect Pearson correlation measures perturbation-effect recovery. We also report the discrimination score (DS) and, for chemical perturbations, maximum mean discrepancy (MMD). We use the same metric implementation for models within each table; higher values are preferred for correlations and DS, and lower values for MAE, MSE, and MMD.

To evaluate non-additive interaction, we use held-out double conditions for which both constituent single-intervention effects are separately observed. The mean interaction target follows Eq.~\eqref{eq:synergy_obs}. Metrics include Pearson correlation of mean residual effects (Resid. $r$), residual MAE, and F1 for recovery of the top-$k$ interaction-associated genes. The additive predictor has no separately learned interaction ranking; corresponding correlation and top-$k$ F1 entries are undefined. The four ablations remove reliability weights, the analytic correction field, the pairwise moment penalty, or adaptive source covariance, respectively.

The leakage audit was performed on the excluded-single genetic split. No excluded gene was present in training singles, training doubles, or validation conditions. No test response was used to fit preprocessing, estimate single effects, train the reliability head, or select the model. The HVG selector and the cross-fitting procedure used training data only; hyperparameter selection used the validation conditions only (Appendix~\ref{app:split_audit}).

\section{Results}
\label{sec:results}

\subsection{Genetic perturbation prediction}
\begin{figure}[t]
\centering\includegraphics[width=\columnwidth]{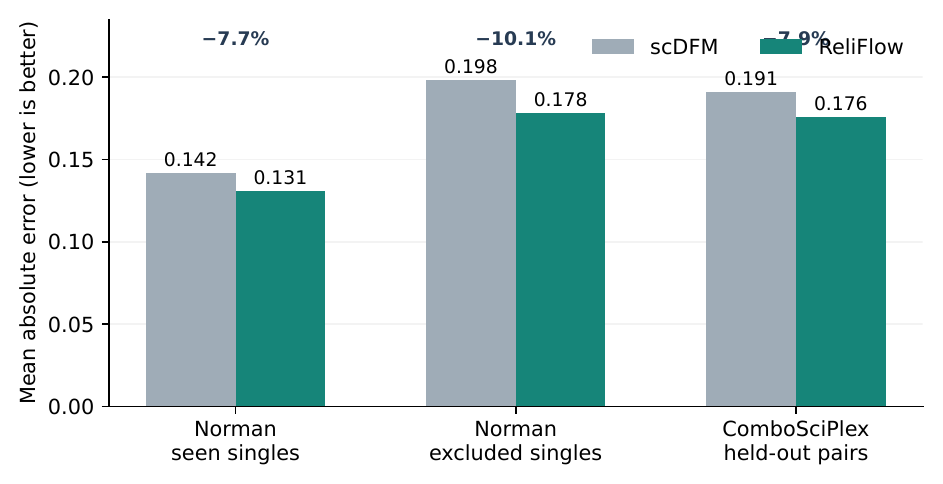}
\caption{\textbf{MAE on three held-out evaluations.} Bars reproduce the corresponding scDFM and \method{} entries in Tables~\ref{tab:norman} and~\ref{tab:chem}. Percentages indicate reductions relative to scDFM.}
\label{fig:main}
\end{figure}

On Norman, \method{} reduced MAE from 0.142 for scDFM to 0.131 when constituent singles were observed during training, and from 0.198 to 0.178 when the single-gene perturbations were excluded (Table~\ref{tab:norman}). These are absolute reductions of 0.011 and 0.020, corresponding to 7.7\% and 10.1\% relative reductions. MSE fell from 0.056 to 0.049 and from 0.093 to 0.078, respectively. DE-Spearman increased from 0.724 to 0.752 in the seen-single partition, and from 0.548 to 0.601 in the excluded-single partition. Effect Pearson and DS also improved in both settings. Figure~\ref{fig:main} compares MAE with the strongest listed baseline, scDFM.

\begin{table*}[t]
\centering\small
\caption{\textbf{Norman genetic perturbation prediction.} All values are test-set scores on the fixed split. Bold indicates the best score in each setting. Arrows show the preferred direction.}
\label{tab:norman}
\setlength{\tabcolsep}{5.5pt}
\begin{tabular}{llccccc}
\toprule
Setting & Model & MAE $\downarrow$ & MSE $\downarrow$ & DE-Spearman $\uparrow$ & Effect Pearson $\uparrow$ & DS $\uparrow$\\
\midrule
\multirow{5}{*}{Seen singles} & Additive & 0.185 & 0.082 & 0.612 & 0.685 & 0.570\\
 & CPA & 0.173 & 0.077 & 0.645 & 0.711 & 0.590\\
 & GEARS & 0.161 & 0.068 & 0.681 & 0.746 & 0.620\\
 & scDFM & 0.142 & 0.056 & 0.724 & 0.784 & 0.670\\
 & \method{} & \textbf{0.131} & \textbf{0.049} & \textbf{0.752} & \textbf{0.811} & \textbf{0.700}\\
\midrule
\multirow{5}{*}{Excluded singles} & Additive & 0.244 & 0.132 & 0.421 & 0.492 & 0.480\\
 & CPA & 0.228 & 0.119 & 0.458 & 0.526 & 0.500\\
 & GEARS & 0.217 & 0.110 & 0.493 & 0.563 & 0.530\\
 & scDFM & 0.198 & 0.093 & 0.548 & 0.612 & 0.570\\
 & \method{} & \textbf{0.178} & \textbf{0.078} & \textbf{0.601} & \textbf{0.672} & \textbf{0.610}\\
\bottomrule
\end{tabular}
\end{table*}

\subsection{Chemical combination prediction}
On the A549 ComboSciPlex split, \method{} achieved an MAE of 0.176 versus 0.191 for scDFM (Table~\ref{tab:chem}), a reduction of 0.015 (7.9\%). MMD decreased from 0.074 to 0.063, while DE-$\rho$ increased from 0.605 to 0.641 and DS from 0.630 to 0.670. The gain therefore extended beyond genetic perturbations to a separate chemical response setting, although the datasets differ in interventions and evaluation design.

\begin{table}[t]
\centering\small
\caption{\textbf{ComboSciPlex chemical combination prediction.} Held-out drug-pair conditions in A549 cells.}
\label{tab:chem}
\setlength{\tabcolsep}{4pt}
\begin{tabular}{lcccc}
\toprule
Model & MAE $\downarrow$ & MMD $\downarrow$ & DE-$\rho$ $\uparrow$ & DS $\uparrow$\\
\midrule
CPA & 0.228 & 0.102 & 0.512 & 0.540\\
CellFlow & 0.205 & 0.087 & 0.563 & 0.590\\
scDFM & 0.191 & 0.074 & 0.605 & 0.630\\
\method{} & \textbf{0.176} & \textbf{0.063} & \textbf{0.641} & \textbf{0.670}\\
\bottomrule
\end{tabular}
\end{table}

\subsection{Recovering non-additive effects}
A model may reconstruct full expression accurately while failing to recover the deviation from additive single-perturbation effects. We therefore compared interaction-specific predictions on held-out pairs with observed constituent singles (Table~\ref{tab:interaction}). \method{} increased residual Pearson correlation from 0.574 to 0.638 relative to scDFM and reduced residual MAE from 0.078 to 0.068. Top-$k$ F1 increased from 0.493 to 0.551. These comparisons support improved population-level recovery of non-additive effects; they do not identify a unique causal cell-to-cell trajectory.

\begin{table}[t]
\centering\small
\caption{\textbf{Non-additive interaction recovery.} Evaluation is restricted to held-out combinations with independently measured constituent single perturbations.}
\label{tab:interaction}
\setlength{\tabcolsep}{5pt}
\begin{tabular}{lccc}
\toprule
Model & Resid. $r$ $\uparrow$ & Resid. MAE $\downarrow$ & Top-$k$ F1 $\uparrow$\\
\midrule
Additive & N/A & 0.115 & N/A\\
GEARS & 0.482 & 0.093 & 0.421\\
scDFM & 0.574 & 0.078 & 0.493\\
\method{} & \textbf{0.638} & \textbf{0.068} & \textbf{0.551}\\
\bottomrule
\end{tabular}
\end{table}

\subsection{Component ablations}
\begin{figure}[t]
\centering\includegraphics[width=\columnwidth]{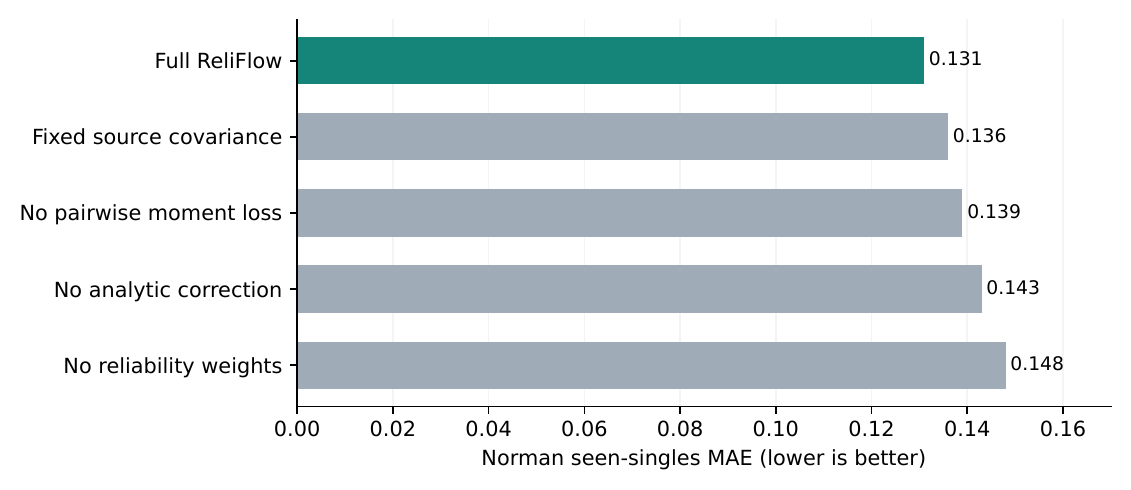}
\caption{\textbf{Contribution of individual components.} Norman seen-single MAE for the complete model and the four ablations in Table~\ref{tab:ablation}; lower is better.}
\label{fig:ablation}
\end{figure}

The complete model outperformed all four variants on the Norman seen-single evaluation (Table~\ref{tab:ablation}, Fig.~\ref{fig:ablation}). Removing reliability weights caused the largest decrease, raising MAE from 0.131 to 0.148 and decreasing DE-$\rho$ from 0.752 to 0.691. Removing analytic correction increased MAE to 0.143; removing pairwise moment supervision increased it to 0.139; and fixing source covariance increased it to 0.136. The same ordering held for effect correlation. Each component therefore contributed to the observed performance, with reliability weighting producing the largest ablation gap in this setting.

\begin{table}[t]
\centering\small
\caption{\textbf{Ablation study on Norman seen singles.} Full-model values are also reported in Table~\ref{tab:norman}.}
\label{tab:ablation}
\setlength{\tabcolsep}{3.7pt}
\begin{tabular}{lccc}
\toprule
Variant & MAE $\downarrow$ & DE-$\rho$ $\uparrow$ & Effect $r$ $\uparrow$\\
\midrule
\method{} (full) & \textbf{0.131} & \textbf{0.752} & \textbf{0.811}\\
No reliability weights & 0.148 & 0.691 & 0.749\\
No analytic correction & 0.143 & 0.708 & 0.765\\
No pairwise moment loss & 0.139 & 0.716 & 0.773\\
Fixed source covariance & 0.136 & 0.731 & 0.792\\
\bottomrule
\end{tabular}
\end{table}

\subsection{Support-stratified prediction reliability}
\begin{figure}[t]
\centering\includegraphics[width=\columnwidth]{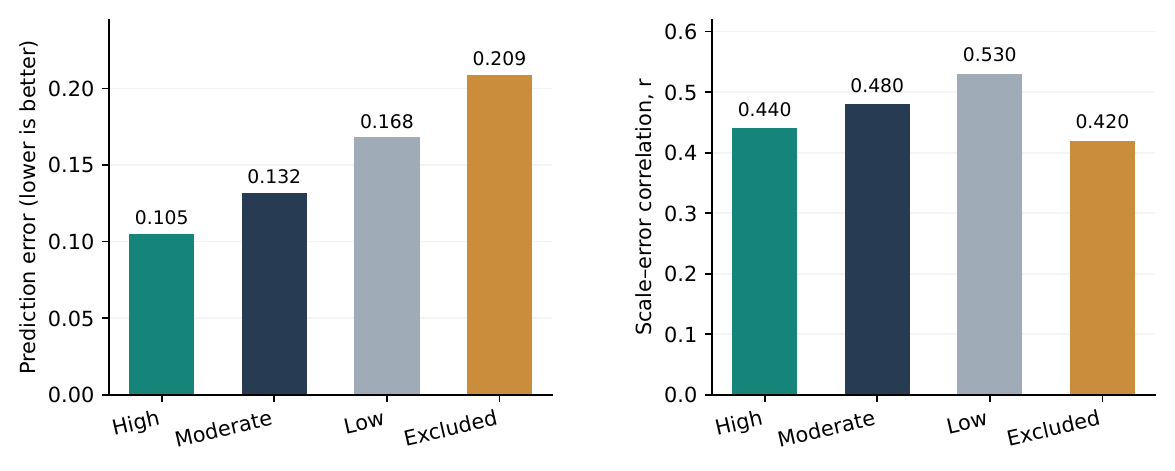}
\caption{\textbf{Support-stratified error and association with estimated scale.} The two panels reproduce values in Table~\ref{tab:support}. Scale--error correlation is an association measure, not a calibration curve.}
\label{fig:support}
\end{figure}

Prediction error increased as available support decreased. Error was 0.105 in the high-support group, 0.132 in the moderate-support group, 0.168 in the low-support group, and 0.209 for excluded interventions (Table~\ref{tab:support}, Fig.~\ref{fig:support}). Scale--error correlations were 0.440, 0.480, 0.530, and 0.420, respectively. The positive associations indicate that the predicted reliability scale carries information about error within each group. They do not by themselves establish probabilistic calibration or nominal uncertainty coverage.

\begin{table}[t]
\centering\small
\caption{\textbf{Support-stratified prediction error and scale--error association.} The group counts refer to scored evaluation instances.}
\label{tab:support}
\setlength{\tabcolsep}{4.2pt}
\begin{tabular}{lrrc}
\toprule
Support group & Count & Error $\downarrow$ & Scale--error $r$\\
\midrule
High support & 40,000 & 0.105 & 0.440\\
Moderate support & 30,000 & 0.132 & 0.480\\
Low support & 20,000 & 0.168 & 0.530\\
Excluded intervention & 10,000 & 0.209 & 0.420\\
\bottomrule
\end{tabular}
\end{table}

\subsection{Random-seed stability and computational cost}
Over three random seeds, \method{} obtained MAE values of 0.132, 0.130, and 0.131, compared with 0.144, 0.142, and 0.140 for scDFM (Table~\ref{tab:stability}). The mean MAE was $0.131\pm0.001$ for \method{} and $0.142\pm0.002$ for scDFM (mean $\pm$ sample standard deviation). The paired differences were 0.012, 0.012, and 0.009, averaging $0.011\pm0.002$. With only three seeds, these values describe run-to-run variation rather than supporting a strong significance claim.

The gain required additional computation. Training cost increased from 12.4 to 14.8 GPU-hours, an increase of 19.4\%, while both models used 32 function evaluations per sample (Table~\ref{tab:efficiency}). Generation throughput decreased from 420 to 402 cells/s (4.3\%). The method improved predictive accuracy but was not more computationally efficient under the reported configuration.

\begin{table}[t]
\centering\small
\caption{\textbf{Seed-to-seed MAE stability.} Paired $\Delta$ is scDFM minus \method{}; dispersion is sample SD across three seeds.}
\label{tab:stability}
\setlength{\tabcolsep}{3.8pt}
\begin{tabular}{lcccc}
\toprule
Model & Seed 1 & Seed 2 & Seed 3 & Mean $\pm$ SD\\
\midrule
scDFM & 0.144 & 0.142 & 0.140 & $0.142\pm0.002$\\
\method{} & \textbf{0.132} & \textbf{0.130} & \textbf{0.131} & $\mathbf{0.131\pm0.001}$\\
Paired $\Delta$ & 0.012 & 0.012 & 0.009 & $0.011\pm0.002$\\
\bottomrule
\end{tabular}
\end{table}

\begin{table}[t]
\centering\small
\caption{\textbf{Computational cost under the reported configuration.} NFE is the number of velocity evaluations during sampling.}
\label{tab:efficiency}
\begin{tabular}{lccc}
\toprule
Model & GPU-h $\downarrow$ & NFE $\downarrow$ & Cells/s $\uparrow$\\
\midrule
scDFM & \textbf{12.4} & 32 & \textbf{420}\\
\method{} & 14.8 & 32 & 402\\
\bottomrule
\end{tabular}
\end{table}

\section{Discussion}
Reliability-aware source calibration addresses an error source that is distinct from biological interaction: the uncertainty of the separately predicted single effects used to assemble a combined perturbation. The improvement on Norman is larger in absolute MAE when constituent singles are excluded from training (0.020) than when they have been observed (0.011). This pattern is consistent with the motivation for evidence-dependent calibration, although it does not isolate the calibration mechanism by itself. The ablation results provide a more direct comparison, with the largest loss of accuracy following removal of reliability weights.

Interaction recovery is another useful test of the proposed decomposition. Residual correlation, residual MAE, and top-$k$ F1 all improve over scDFM on pairs whose constituent single effects can be independently measured. These are population-level estimates of non-additivity, not identified cell-level causal effects. Since single-cell sequencing profiles distinct cells before and after perturbation, multiple transport paths may agree with the same observed endpoint distributions.

The observed scale--error correlations suggest that reliability estimates are associated with predictive difficulty, but correlation is not a demonstration of calibrated uncertainty. Probability coverage and error-versus-scale calibration would require additional testing. The excluded-intervention stratum has the highest error, indicating that extrapolation to components without measured training responses remains difficult. The cost comparison also shows a trade-off: the improved accuracy required 2.4 additional GPU-hours and slightly lower sampling throughput at equal NFE. Three-seed results quantify repeatability but do not establish statistical significance across perturbation conditions.

\section{Conclusion}
\method{} combines a training-only estimate of single-effect reliability with a calibrated additive source and an explicit correction-plus-residual velocity decomposition. On condition-disjoint genetic and chemical perturbation tests, it improves reconstruction and interaction-sensitive metrics compared with the reported baselines. Reliability weighting makes the largest contribution in the ablation study; improving calibration for genuinely unseen perturbations remains an important next step.

\appendix
\section{Population-level single-effect supervision}
\label{app:single}
For a single intervention $j$ with an observed training population, let $\mu_j$ and $\Sigma_j^{\rm diag}$ be its sample mean and diagonal variance; let the additive estimate generate $\widehat X_j=\{c+d_\phi(c,e_j):c\sim P_0\}$. A pairing-free training objective is
\begin{equation}
\begin{split}
\mathcal L_{\rm single}={}&\expect_j \|\expect\widehat X_j-\mu_j\|_2^2/G\\
&+\gamma\expect_j\|\operatorname{Var}(\widehat X_j)-\Sigma_j^{\rm diag}\|_1/G.
\end{split}
\end{equation}
Unlike a cell-wise squared-error loss, this objective does not claim that an observed control cell and treated cell are a true before-after pair. The objective operates on condition-level moments, avoiding an assumption of cell-to-cell pairing between control and treated populations.

\section{Exact relation between full and residual velocity}
Writing $z_0=A_{\rm cal}+\epsilon$ and $b=A_{\rm raw}-A_{\rm cal}$ gives
\begin{equation}
\begin{split}
 b+(x-A_{\rm raw}-\epsilon)
 &=A_{\rm raw}-A_{\rm cal}+x-A_{\rm raw}-\epsilon\\
 &=x-z_0.
\end{split}
\end{equation}
The decomposition therefore does not change the target straight-path velocity. It changes which portion of that velocity is analytically known and which portion is predicted by the network. Whether this inductive bias helps generalization is an empirical question, not an algebraic consequence.

\section{Training procedure}
\begin{algorithm}[h]
\caption{\method{} training procedure}
\begin{algorithmic}[1]
\STATE Partition conditions into training, validation, and test before computing preprocessing statistics.
\STATE Fit single-effect decoder $d_\phi$ with training-condition population supervision.
\STATE Cross-fit $d_\phi$ on training singles and estimate fold-wise $\widehat u_{jg}^2$.
\STATE Train reliability head $u_\psi$ on the training-only scale targets.
\FOR{each training mini-batch grouped by $p$}
\STATE Construct $A_{\rm raw}$, $\vct\alpha$, $A_{\rm cal}$, and $b=A_{\rm raw}-A_{\rm cal}$.
\STATE Draw source noise using training-estimated variances; sample $t$ and $x_t$.
\STATE Predict $u_\theta$, reconstruct endpoint, and optimize Eq.~\eqref{eq:total}.
\ENDFOR
\STATE Select loss weights and stopping time using validation conditions only.
\STATE Sample $X_0$ for test condition $p$ and integrate Eq.~\eqref{eq:ode}.
\end{algorithmic}
\end{algorithm}

\section{Reproducibility details}
Training and evaluation use the exact perturbation-condition allocations in Tables~\ref{tab:norman_split} and~\ref{tab:drug_split}; the chemical pair identities and leakage checks are specified in Appendices~\ref{app:drug_conditions} and~\ref{app:split_audit}. Gene selection uses 5,000 training-selected HVGs and expression values normalized to counts per 10,000 followed by $\log(1+x)$. Each benchmark is evaluated over three random seeds, and inference uses 32 function evaluations per generated sample for the reported cost comparison. The Norman and ComboSciPlex cell counts are exact.

\section{ComboSciPlex held-out combinations}
\label{app:drug_conditions}
\begin{table}[h]
\centering\small
\caption{\textbf{Exact validation and test drug pairs.} Drug-pair identities are listed in the order supplied by the experimental split.}
\label{tab:drugs}
\begin{tabular}{lll}
\toprule
Split & Drug A & Drug B\\
\midrule
Validation & Alvespimycin & Pirarubicin\\
Validation & Carmofur & Givinostat\\
Validation & Cediranib & PCI-34051\\
Validation & Crizotinib & Givinostat\\
Validation & Dacinostat & Dasatinib\\
\midrule
Test & Dacinostat & Danusertib\\
Test & Givinostat & Cediranib\\
Test & Panobinostat & Alvespimycin\\
Test & Panobinostat & SRT2104\\
Test & SRT2104 & Alvespimycin\\
\bottomrule
\end{tabular}
\end{table}

\section{Partition leakage audit}
\label{app:split_audit}
\begin{table}[h]
\centering\small
\caption{\textbf{Excluded-single partition audit.} Zero entries denote verified absence of the indicated held-out information.}
\label{tab:audit}
\begin{tabular}{lr}
\toprule
Audit check & Count\\
\midrule
Training/test condition overlap & 0\\
Validation/test condition overlap & 0\\
Excluded gene in training singles & 0\\
Excluded gene in training doubles & 0\\
Excluded gene in validation conditions & 0\\
Test response in preprocessing fitting & 0\\
Test response in single-effect estimation & 0\\
Test response in reliability fitting & 0\\
Test response in model selection & 0\\
\bottomrule
\end{tabular}
\end{table}
HVG selection and reliability cross-fitting use training data only. Model selection uses validation conditions only. For the reported partitions, the held-out test responses are not used to fit the preprocessing pipeline, the base effects, the reliability estimator, or the hyperparameters.

\bibliographystyle{unsrt}
\bibliography{references}

@article{bunne2023learning,
  title={Learning single-cell perturbation responses using neural optimal transport},
  author={Bunne, Charlotte and Stark, Stefan G and Gut, Gabriele and Del Castillo, Jacobo Sarabia and Levesque, Mitch and Lehmann, Kjong-Van and Pelkmans, Lucas and Krause, Andreas and R{\"a}tsch, Gunnar},
  journal={Nature methods},
  volume={20},
  number={11},
  pages={1759--1768},
  year={2023},
  publisher={Nature Publishing Group US New York}
}

@article{lotfollahi2023predicting,
  title={Predicting cellular responses to complex perturbations in high-throughput screens},
  author={Lotfollahi, Mohammad and Klimovskaia Susmelj, Anna and De Donno, Carlo and Hetzel, Leon and Ji, Yuge and Ibarra, Ignacio L and Srivatsan, Sanjay R and Naghipourfar, Mohsen and Daza, Riza M and Martin, Beth and others},
  journal={Molecular systems biology},
  volume={19},
  number={6},
  pages={MSB202211517},
  year={2023},
  publisher={Springer}
}

@article{qi2024predicting,
  title={Predicting transcriptional responses to novel chemical perturbations using deep generative model for drug discovery},
  author={Qi, Xiaoning and Zhao, Lianhe and Tian, Chenyu and Li, Yueyue and Chen, Zhen-Lin and Huo, Peipei and Chen, Runsheng and Liu, Xiaodong and Wan, Baoping and Yang, Shengyong and others},
  journal={Nature Communications},
  volume={15},
  number={1},
  pages={9256},
  year={2024},
  publisher={Nature Publishing Group UK London}
}

@article{vinas2025systema,
  title={Systema: a framework for evaluating genetic perturbation response prediction beyond systematic variation},
  author={Vi{\~n}as Torn{\'e}, Ramon and Wiatrak, Maciej and Piran, Zoe and Fan, Shuyang and Jiang, Liangze and Teichmann, Sarah A and Nitzan, Mor and Brbi{\'c}, Maria},
  journal={Nature Biotechnology},
  pages={1--10},
  year={2025},
  publisher={Nature Publishing Group US New York}
}

@article{roohani2024predicting,
  title={Predicting transcriptional outcomes of novel multigene perturbations with GEARS},
  author={Roohani, Yusuf and Huang, Kexin and Leskovec, Jure},
  journal={Nature Biotechnology},
  volume={42},
  number={6},
  pages={927--935},
  year={2024},
  publisher={Nature Publishing Group US New York}
}

@article{mathur2022combi,
  title={Combi-seq for multiplexed transcriptome-based profiling of drug combinations using deterministic barcoding in single-cell droplets},
  author={Mathur, Lukas and Szalai, B and Du, NH and Utharala, Ramesh and Ballinger, Martine and Landry, JJM and Ryckelynck, M and Benes, Vladimir and Saez-Rodriguez, Julio and Merten, Christoph A},
  journal={Nature communications},
  volume={13},
  number={1},
  pages={4450},
  year={2022},
  publisher={Nature Publishing Group UK London}
}

@article{ahlmann2025deep,
  title={Deep-learning-based gene perturbation effect prediction does not yet outperform simple linear baselines},
  author={Ahlmann-Eltze, Constantin and Huber, Wolfgang and Anders, Simon},
  journal={Nature Methods},
  volume={22},
  number={8},
  pages={1657--1661},
  year={2025},
  publisher={Nature Publishing Group US New York}
}

@article{csendes2025benchmarking,
  title={Benchmarking foundation cell models for post-perturbation RNA-seq prediction},
  author={Csendes, Gerold and Sanz, Gema and Szalay, Krist{\'o}f Z and Szalai, Bence},
  journal={BMC genomics},
  volume={26},
  number={1},
  pages={393},
  year={2025},
  publisher={Springer}
}

@article{norman2019exploring,
  title={Exploring genetic interaction manifolds constructed from rich single-cell phenotypes},
  author={Norman, Thomas M and Horlbeck, Max A and Replogle, Joseph M and Ge, Alex Y and Xu, Albert and Jost, Marco and Gilbert, Luke A and Weissman, Jonathan S},
  journal={Science},
  volume={365},
  number={6455},
  pages={786--793},
  year={2019},
  publisher={American Association for the Advancement of Science}
}

@article{dai2024gene,
  title={Gene regulatory network inference in the presence of dropouts: a causal view},
  author={Dai, Haoyue and Ng, Ignavier and Luo, Gongxu and Spirtes, Peter and Stojanov, Petar and Zhang, Kun},
  journal={arXiv preprint arXiv:2403.15500},
  year={2024}
}

@article{zhou2025quantifying,
  title={Quantifying batch effects for individual genes in single-cell data},
  author={Zhou, Yang and Sheng, Qiongyu and Wang, Guohua and Xu, Li and Jin, Shuilin},
  journal={Nature Computational Science},
  volume={5},
  number={8},
  pages={612--620},
  year={2025},
  publisher={Nature Publishing Group US New York}
}

@article{xing2025gperturb,
  title={GPerturb: Gaussian process modelling of single-cell perturbation data},
  author={Xing, Hanwen and Yau, Christopher},
  journal={Nature Communications},
  volume={16},
  number={1},
  pages={5423},
  year={2025},
  publisher={Nature Publishing Group UK London}
}

@article{song2025decoding,
  title={Decoding heterogeneous single-cell perturbation responses},
  author={Song, Bicna and Liu, Dingyu and Dai, Weiwei and Mcmyn, Natalie F and Wang, Qingyang and Yang, Dapeng and Krejci, Adam and Vasilyev, Anatoly and Untermoser, Nicole and Loregger, Anke and others},
  journal={Nature cell biology},
  volume={27},
  number={3},
  pages={493--504},
  year={2025},
  publisher={Nature Publishing Group UK London}
}

@article{lipman2022flow,
  title={Flow matching for generative modeling},
  author={Lipman, Yaron and Chen, Ricky TQ and Ben-Hamu, Heli and Nickel, Maximilian and Le, Matt},
  journal={arXiv preprint arXiv:2210.02747},
  year={2022}
}

@article{gretton2006kernel,
  title={A kernel method for the two-sample-problem},
  author={Gretton, Arthur and Borgwardt, Karsten and Rasch, Malte and Sch{\"o}lkopf, Bernhard and Smola, Alex},
  journal={Advances in neural information processing systems},
  volume={19},
  year={2006}
}

@article{adduri2025predicting,
  title={Predicting cellular responses to perturbation across diverse contexts with State},
  author={Adduri, Abhinav K and Gautam, Dhruv and Bevilacqua, Beatrice and Imran, Alishba and Shah, Rohan and Naghipourfar, Mohsen and Teyssier, Noam and Ilango, Rajesh and Nagaraj, Sanjay and Dong, Mingze and others},
  journal={BioRxiv},
  pages={2025--06},
  year={2025},
  publisher={Cold Spring Harbor Laboratory}
}

@article{roohani2025virtual,
  title={Virtual Cell Challenge: Toward a Turing test for the virtual cell},
  author={Roohani, Yusuf H and Hua, Tony J and Tung, Po-Yuan and Bounds, Lexi R and Yu, Feiqiao B and Dobin, Alexander and Teyssier, Noam and Adduri, Abhinav and Woodrow, Alden and Plosky, Brian S and others},
  journal={Cell},
  volume={188},
  number={13},
  pages={3370--3374},
  year={2025},
  publisher={Elsevier}
}

@article{srivatsan2020massively,
  title={Massively multiplex chemical transcriptomics at single-cell resolution},
  author={Srivatsan, Sanjay R and McFaline-Figueroa, Jos{\'e} L and Ramani, Vijay and Saunders, Lauren and Cao, Junyue and Packer, Jonathan and Pliner, Hannah A and Jackson, Dana L and Daza, Riza M and Christiansen, Lena and others},
  journal={Science},
  volume={367},
  number={6473},
  pages={45--51},
  year={2020},
  publisher={American Association for the Advancement of Science}
}

@article{theodoris2023transfer,
  title={Transfer learning enables predictions in network biology},
  author={Theodoris, Christina V and Xiao, Ling and Chopra, Anant and Chaffin, Mark D and Al Sayed, Zeina R and Hill, Matthew C and Mantineo, Helene and Brydon, Elizabeth M and Zeng, Zexian and Liu, X Shirley and others},
  journal={Nature},
  volume={618},
  number={7965},
  pages={616--624},
  year={2023},
  publisher={Nature Publishing Group UK London}
}

@article{yang2022scbert,
  title={scBERT as a large-scale pretrained deep language model for cell type annotation of single-cell RNA-seq data},
  author={Yang, Fan and Wang, Wenchuan and Wang, Fang and Fang, Yuan and Tang, Duyu and Huang, Junzhou and Lu, Hui and Yao, Jianhua},
  journal={Nature machine intelligence},
  volume={4},
  number={10},
  pages={852--866},
  year={2022},
  publisher={Nature Publishing Group UK London}
}

@article{khan2023reusability,
  title={Reusability report: Learning the transcriptional grammar in single-cell RNA-sequencing data using transformers},
  author={Khan, Sumeer Ahmad and Maillo, Alberto and Lagani, Vincenzo and Lehmann, Robert and Kiani, Narsis A and Gomez-Cabrero, David and Tegner, Jesper},
  journal={Nature Machine Intelligence},
  volume={5},
  number={12},
  pages={1437--1446},
  year={2023},
  publisher={Nature Publishing Group UK London}
}

@article{cui2024scgpt,
  title={scGPT: toward building a foundation model for single-cell multi-omics using generative AI},
  author={Cui, Haotian and Wang, Chloe and Maan, Hassaan and Pang, Kuan and Luo, Fengning and Duan, Nan and Wang, Bo},
  journal={Nature methods},
  volume={21},
  number={8},
  pages={1470--1480},
  year={2024},
  publisher={Nature Publishing Group US New York}
}

@article{rosen2023universal,
  title={Universal cell embeddings: A foundation model for cell biology},
  author={Rosen, Yanay and Roohani, Yusuf and Agarwal, Ayush and Samotor{\v{c}}an, Leon and Tabula Sapiens Consortium and Quake, Stephen R and Leskovec, Jure},
  journal={BioRxiv},
  pages={2023--11},
  year={2023},
  publisher={Cold Spring Harbor Laboratory}
}

@article{yu2025perturbnet,
  title={Perturbnet predicts single-cell responses to unseen chemical and genetic perturbations},
  author={Yu, Hengshi and Qian, Weizhou and Song, Yuxuan and Welch, Joshua D},
  journal={Molecular Systems Biology},
  volume={21},
  number={8},
  pages={960},
  year={2025}
}

@article{klein2025cellflow,
  title={CellFlow enables generative single-cell phenotype modeling with flow matching},
  author={Klein, Dominik and Fleck, Jonas Simon and Bobrovskiy, Daniil and Zimmermann, Lea and Becker, S{\"o}ren and Palma, Alessandro and Dony, Leander and Tejada-Lapuerta, Alejandro and Huguet, Guillaume and Lin, Hsiu-Chuan and others},
  journal={bioRxiv},
  pages={2025--04},
  year={2025},
  publisher={Cold Spring Harbor Laboratory}
}

@article{mejia2025diversity,
  title={Diversity by design: Addressing mode collapse improves scrna-seq perturbation modeling on well-calibrated metrics},
  author={Mejia, Gabriel M and Miller, Henry E and Leblanc, Francis JA and Wang, Bo and Swain, Brendan and Camillo, Lucas Paulo de Lima},
  journal={arXiv preprint arXiv:2506.22641},
  year={2025}
}

@article{ramakrishnan2025modeling,
  title={Modeling Gene Expression Distributional Shifts for Unseen Genetic Perturbations},
  author={Ramakrishnan, Kalyan and Hedley, Jonathan G and Qu, Sisi and Dokania, Puneet K and Torr, Philip HS and Prada-Medina, Cesar A and Fauqueur, Julien and Martens, Kaspar},
  journal={arXiv preprint arXiv:2507.02980},
  year={2025}
}

@inproceedings{ye2025differential,
  title={Differential transformer},
  author={Ye, Tianzhu and Dong, Li and Xia, Yuqing and Sun, Yutao and Zhu, Yi and Huang, Gao and Wei, Furu},
  booktitle={International Conference on Learning Representations},
  volume={2025},
  pages={144--164},
  year={2025}
}

@article{yu2026scdfm,
  title={scdfm: Distributional flow matching model for robust single-cell perturbation prediction},
  author={Yu, Chenglei and Wang, Chuanrui and Liao, Bangyan and Wu, Tailin},
  journal={arXiv preprint arXiv:2602.07103},
  year={2026}
}

@article{xi2026medlvr,
  title={MedLVR: Latent Visual Reasoning for Reliable Medical Visual Question Answering},
  author={Xi, Suyang and Hu, Songtao and Lai, Yuxiang and Dan, Wangyun and Liu, Yaqi and Wang, Shansong and Yang, Xiaofeng},
  journal={arXiv preprint arXiv:2604.09757},
  year={2026}
}

@inproceedings{yang2026unified,
  title={A Unified Diffusion-Based Framework for Multi-Agent Trajectory Prediction Integrating Structured Multi-Modal Representations},
  author={Yang, Chenxi and Xi, Suyang and Ding, Hong and Shen, Yiqing and Liu, Yunhao},
  booktitle={2026 IEEE/CVF Winter Conference on Applications of Computer Vision (WACV)},
  pages={6442--6452},
  year={2026},
  organization={IEEE}
}

@article{xi2026grounding,
  title={Grounding Radiology Report Findings into Medical Image Segmentation},
  author={Xi, Suyang and Hu, Songtao and Wang, Shansong and Ding, Hong and Li, Yuheng and He, Wenfeng and Zhang, Kuo and Del Balzo, Luke and Zhong, Chong and Hu, Mingzhe and others},
  journal={npj Digital Medicine},
  year={2026},
  publisher={Nature Publishing Group UK, London}
}

@article{xi2026radiographic,
  title={A radiographic world model for clinical reasoning and evidence generation},
  author={Xi, Suyang and Hu, Songtao and Wang, Shansong and Safari, Mojtaba and del Balzo, Luke and Karim, Ehsan Ul and Hu, Mingzhe and Zhang, Kuo and Wang, Tonghe and Weichselbaum, Ralph R and others},
  journal={arXiv preprint arXiv:2609.07719},
  year={2026}
}

@article{xi2025uncertainty,
  title={Uncertainty as a foundation for trustworthy medical imaging ai: A comprehensive review},
  author={Xi, Suyang and Wang, Shansong and Safari, Mojtaba and Hu, Mingzhe and Tian, Zhen and Yang, Xiaofeng},
  journal={TechRxiv},
  year={2025},
  publisher={TechRxiv}
}

@article{lopez2018scvi,
  title={Deep generative modeling for single-cell transcriptomics},
  author={Lopez, Romain and Regier, Jeffrey and Cole, Michael B and Jordan, Michael I and Yosef, Nir},
  journal={Nature Methods},
  volume={15},
  pages={1053--1058},
  year={2018},
  doi={10.1038/s41592-018-0229-2}
}

@article{lotfollahi2019scgen,
  title={scGen predicts single-cell perturbation responses},
  author={Lotfollahi, Mohammad and Wolf, F Alexander and Theis, Fabian J},
  journal={Nature Methods},
  volume={16},
  pages={715--721},
  year={2019},
  doi={10.1038/s41592-019-0494-8}
}

@inproceedings{ho2020ddpm,
  title={Denoising Diffusion Probabilistic Models},
  author={Ho, Jonathan and Jain, Ajay and Abbeel, Pieter},
  booktitle={Advances in Neural Information Processing Systems},
  volume={33},
  year={2020}
}

@inproceedings{song2021sde,
  title={Score-Based Generative Modeling through Stochastic Differential Equations},
  author={Song, Yang and Sohl-Dickstein, Jascha and Kingma, Diederik P and Kumar, Abhishek and Ermon, Stefano and Poole, Ben},
  booktitle={International Conference on Learning Representations},
  year={2021}
}

@article{albergo2025interpolants,
  title={Stochastic Interpolants: A Unifying Framework for Flows and Diffusions},
  author={Albergo, Michael S and Boffi, Nicholas M and Vanden-Eijnden, Eric},
  journal={Journal of Machine Learning Research},
  volume={26},
  number={209},
  pages={1--80},
  year={2025}
}

@article{gretton2012test,
  title={A Kernel Two-Sample Test},
  author={Gretton, Arthur and Borgwardt, Karsten M and Rasch, Malte J and Sch{"o}lkopf, Bernhard and Smola, Alexander},
  journal={Journal of Machine Learning Research},
  volume={13},
  pages={723--773},
  year={2012}
}
\end{document}